\documentclass{article}

\usepackage{arxiv}

\usepackage[utf8]{inputenc} % allow utf-8 input
\usepackage[T1]{fontenc}    % use 8-bit T1 fonts
\usepackage{hyperref}       % hyperlinks
\usepackage{url}            % simple URL typesetting
\usepackage{booktabs}       % professional-quality tables
\usepackage{amsfonts}       % blackboard math symbols
\usepackage{nicefrac}       % compact symbols for 1/2, etc.
\usepackage{microtype}      % microtypography
\usepackage{amsmath}
\usepackage{cleveref}       % smart cross-referencing
\usepackage{lipsum}         % Can be removed after putting your text content
\usepackage{graphicx}
\usepackage{natbib}
\usepackage{doi}
\usepackage{wrapfig}
\usepackage{adjustbox}

\title{Adaptive Aerial Loco-Manipulation Policy Learning via LLM-Assisted Proactive Task Generation}

\newif\ifuniqueAffiliation
\uniqueAffiliationfalse

\ifuniqueAffiliation % Standard variant of author block
\author{ \href{https://orcid.org/0000-0000-0000-0000}{\includegraphics[scale=0.06]{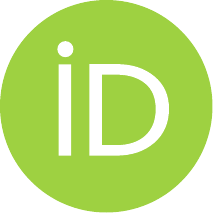}\hspace{1mm}David S.~Hippocampus}\thanks{Use footnote for providing further
		information about author (webpage, alternative
		address)---\emph{not} for acknowledging funding agencies.} \\
	Department of Computer Science\\
	Cranberry-Lemon University\\
	Pittsburgh, PA 15213 \\
	\texttt{hippo@cs.cranberry-lemon.edu} \\
	\And
	\href{https://orcid.org/0000-0000-0000-0000}{\includegraphics[scale=0.06]{orcid.pdf}\hspace{1mm}Elias D.~Striatum} \\
	Department of Electrical Engineering\\
	Mount-Sheikh University\\
	Santa Narimana, Levand \\
	\texttt{stariate@ee.mount-sheikh.edu} \\
}
\else
\usepackage{authblk}

\author[1]{Jingyu Ru\textsuperscript{*}}
\author[1]{Xiangyuan Xie\textsuperscript{*}}
\author[2, 3]{Lixuan Jin}
\author[1]{Jian An}
\author[4]{Xingming Zhang}
\author[1]{Binghao Li}
\author[2, 3]{Siyu Wang}
\author[2, 3]{Xiang Zhou}
\author[2, 3]{Lei Yuan\textsuperscript{$\dagger$}}
\author[1]{Hongli Xu\textsuperscript{$\dagger$}}
\author[2, 3]{Yang Yu}
\affil[1]{Faculty of Robot Science and Engineering, Northeastern University, Shenyang, China}
\affil[2]{National Key Laboratory of Novel Software Technology, Nanjing University, Nanjing, China}
\affil[3]{School of Artificial Intelligence, Nanjing University, Nanjing, China}
\affil[4]{College of Information Science and Engineering, Northeastern University, Shenyang, China}
\fi

\renewcommand{\shorttitle}{ALMA}

\hypersetup{
pdftitle={Adaptive Aerial Loco-Manipulation Policy Learning via LLM-Assisted Proactive Task Generation},
pdfsubject={q-bio.NC, q-bio.QM},
pdfauthor={David S.~Hippocampus, Elias D.~Striatum},
pdfkeywords={First keyword, Second keyword, More},
}

\begin{document}
\maketitle

\begingroup
\renewcommand{\thefootnote}{\fnsymbol{footnote}}
\footnotetext[1]{These authors contributed equally.}
\footnotetext[2]{Corresponding authors: Lei Yuan
(\texttt{
yuanl@lamda.nju.edu.cn}) and Hongli Xu
(\texttt{xuhongli@mail.neu.edu.cn}).}
\endgroup

\begin{abstract}
Loco-manipulation extends the workspace of conventional manipulation by combining mobility with physical interaction, while aerial loco-manipulation further expands this capability into three-dimensional space, enabling robots to reach environments inaccessible to ground-based systems. Moving to an aerial platform, however, gives rise to a demanding control problem, since the vehicle must track trajectories accurately in three dimensions while the motion of the arm and its interaction with the environment continuously shift the center of mass, alter the inertia, and impose force and moment disturbances on the body. Conventional controllers address these effects through explicit models and estimation, whereas learning-based controllers infer dynamics from experience, yet control rewards provide no direct target for the response information retained from history, and fixed task sampling can leave persistent tracking difficulties insufficiently explored. We present \textbf{ALMA}, an aerial loco-manipulation framework for adaptive flight control under external arm commands. ALMA employs a contextual flight policy that adapts from recent measurements and actions, using action-conditioned prediction of future measurement representations to train its context to retain response information relevant to flight control. To expand the experience available for this adaptation, an evolutionary process uses a large language model (LLM) to analyze observed control difficulties and propose new structures of arm motion for subsequent training. In simulation, ALMA reduces position root-mean-square error (RMSE) by up to \textbf{24.7\%} relative to the strongest baseline, while the same policy transfers zero-shot to the aerial manipulator and reduces hover position RMSE by \textbf{79.6\%} during periodic arm motion relative to the real-world baseline.
\end{abstract}

% keywords can be removed
% \keywords{First keyword \and Second keyword \and More}

\section{Introduction}
\label{sec:introduction}

Loco-manipulation extends the workspace of conventional manipulation by combining mobility with physical interaction, while aerial loco-manipulation further expands this capability into three-dimensional space, enabling robots to reach and operate in elevated, cluttered, and hazardous environments inaccessible to ground-based systems~\cite{ollero2022past}. However, changes in arm configuration continuously alter the mass distribution and inertia of the complete system, while joint accelerations and physical contact generate reaction forces and moments on the aerial base~\cite{yadav2024modular}. The flight controller must therefore maintain accurate trajectory tracking while adapting to changes in system response that are only partially revealed by available measurements.

Conventional aerial manipulation controllers compensate for uncertain manipulation effects through analytically designed feedback and adaptive laws~\cite{yadav2024modular,sharma2025impedance}, whereas learning-based methods infer dynamics information from experience. Aco2~\cite{jin2026autonomous} learns a history-dependent context using contrastive supervision to distinguish payload conditions, while the controller in~\cite{ujjawal2026learn} learns multi-step dynamics residuals and adapts their predictions online for disturbance compensation. Their auxiliary objectives respectively organize context by task identity and train a residual predictor for a prescribed compensation law, while predictive-state methods use future observations to supervise the internal representation used for decision making~\cite{venkatraman2017predictive,hefny2018recurrent}. For aerial manipulation, similar instantaneous measurements can correspond to different hidden actuator and disturbance states that require different corrective actions. Predicting future measurement representations conditioned on recorded actions provides a direct training target for retaining these response distinctions in the history representation used by the flight policy. Since the available response information depends on the motion conditions encountered during training, repeatedly sampling fixed motion patterns can leave persistent tracking difficulties insufficiently explored, motivating adaptive sampling~\cite{jiang2021plr,mehta2020adr} and task generation~\cite{parker2022evolving} that direct subsequent experience according to the policy's evolving performance.

We present \textbf{ALMA}, an aerial loco-manipulation framework for adaptive flight control under external arm commands. Inspired by the JEPA~\cite{assran2023ijepa}, ALMA trains its contextual flight policy by predicting learned representations of future measurements from recent history and subsequent actions, thereby supervising the information retained for control. Since learning this adaptation also depends on the responses encountered during training, ALMA uses a large language model (LLM) to analyze episode records, identify motion patterns associated with poor tracking, and propose new structures of arm motion that direct subsequent experience toward difficulties insufficiently addressed by the initial task distribution. Predictive supervision thus trains the context using measured action--response relationships, while task generation broadens the motion conditions that supply subsequent experience, connecting response learning with the policy's remaining tracking difficulties. Both components operate during training, while deployment retains the contextual flight policy to adapt to the changing effects of manipulation.

In simulation, ALMA achieves the lowest mean position RMSE among evaluated controllers, reducing error relative to the strongest baseline by \textbf{24.5\%} under static arm commands and \textbf{24.7\%} under periodic commands, while ablations support the tracking benefit of predictive supervision and the generalization benefit of task generation, with base performance retained. The same policy transfers zero-shot to the aerial manipulator and reduces hover position RMSE by \textbf{79.6\%} during periodic arm motion relative to the real-world baseline.

\section{Related Work}
\label{sec:related_work}

\subsection{Aerial Manipulation}

Aerial manipulation has been studied using model-based planning and feedback control~\cite{tzoumanikas2020aerial}, with adaptive approaches compensating for uncertain manipulation effects through online estimation~\cite{yadav2024modular}. Learning-based methods obtain adaptation information from experience, with Aco2~\cite{jin2026autonomous} combining a history-dependent policy with contrastive supervision for suspended payloads and \cite{ujjawal2026learn} predicting multi-step dynamics residuals and adapting their predictions online for compensation. These methods learn information for adaptation by organizing payload-dependent context or estimating corrections to nominal dynamics. Related predictive-state methods, including PSD~\cite{venkatraman2017predictive} and RPSP~\cite{hefny2018recurrent}, use prediction of future observations to supervise the internal state used for decision making, offering a learning principle for the history representation of a flight controller. Building on predictive-state learning and the JEPA principle of prediction in representation space~\cite{assran2023ijepa}, ALMA trains its contextual flight policy to predict future measurement representations conditioned on recorded actions, linking the history retained for adaptation to subsequent system responses.

\subsection{LLM-Assisted Task Generation}

LLMs have been used to construct robot learning curricula, with CurricuLLM~\cite{ryu2025curricullm} generating task code and evaluating the policies trained from it. Feedback also guides program improvement in ReEvo~\cite{ye2024reevo}, which revises heuristics through reflection on evaluated performance, and FunSearch~\cite{romera2024mathematical}, which combines generated program proposals with external evaluation. These methods demonstrate how language-model proposals can be refined using execution results, complementing policy-driven curricula such as PLR~\cite{jiang2021plr}, which prioritizes existing tasks, and ACCEL~\cite{parker2022evolving}, which introduces new training conditions through environment edits. Drawing on these approaches, ALMA uses the current flight policy's execution records to guide the generation of external arm motions, adapting the training experience as the controller's performance changes.

\section{Background}
\label{sec:problem_formulation}

We consider trajectory tracking for an aerial manipulator under external arm commands, where arm motion and uncertain physical conditions produce system responses that are only partially observed. With arm commands outside the flight policy's action space and access to their future values not assumed, the objective is a single policy that adapts its flight actions to the observed response.

Each admissible task $\zeta\in\mathcal{Z}_{\mathrm{adm}}$ specifies a flight reference and an arm-command process, inducing a partially observable Markov decision process (POMDP) $\mathcal{M}_{\zeta}=(\mathcal{S},\mathcal{A},\mathcal{P}_{\zeta},\mathcal{R}_{\zeta},\Omega,\mathcal{O}_{\zeta},\gamma)$. Here, $\mathcal{S}$ and $\mathcal{A}$ are the state and action spaces, $\mathcal{P}_{\zeta}$ is the transition kernel, $\mathcal{R}_{\zeta}$ is the reward, $\Omega$ and $\mathcal{O}_{\zeta}$ define observations, and $\gamma\in[0,1)$ is the discount factor. The state contains physical and hidden process variables together with the previous action, while observations $\mathbf{o}_t\sim\mathcal{O}_{\zeta}(\cdot\mid\mathbf{s}_t)$ provide reference information and available measurements. Actions depend on the history $\mathcal{H}_t=(\mathbf{o}_0,\mathbf{a}_0,\ldots,\mathbf{o}_{t-1},\mathbf{a}_{t-1},\mathbf{o}_t)$ through $\pi_{\theta}(\mathbf{a}_t\mid\mathcal{H}_t)$.

For task $\zeta$, $J(\theta,\zeta)=\mathbb{E}[\sum_{t=0}^{\infty}\gamma^t r_t]$ is the expected return, where the bounded reward $r_t=\mathcal{R}_{\zeta}(\mathbf{s}_t,\mathbf{a}_t)$ evaluates tracking and control behavior, and the expectation includes initial states, hidden physical conditions, and policy trajectories. At training stage $n$, tasks are sampled from a distribution $\rho_n$ over $\mathcal{Z}_{\mathrm{adm}}$, and policy learning maximizes $J_n(\theta)=\mathbb{E}_{\zeta\sim\rho_n}[J(\theta,\zeta)]$. Since $\rho_n$ determines which arm motions contribute to learning, we also consider updating it according to policy performance, starting from $\rho_0=\rho_{\mathrm{base}}$. Section~\ref{sec:method} specifies the policy representation and training task construction.

\begin{figure*}[t]
    \centering
    \includegraphics[width=1.0\linewidth]{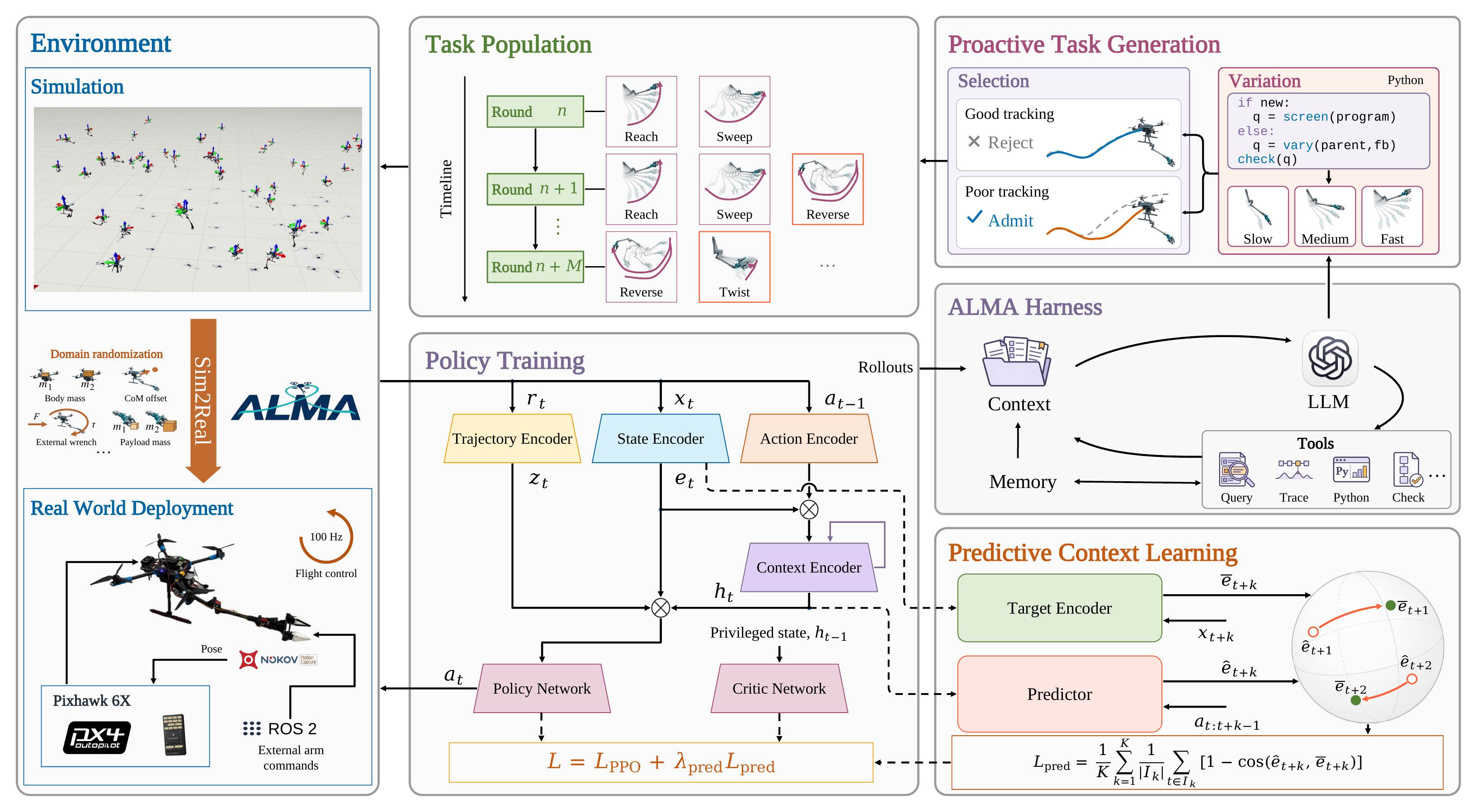}
    \caption{Overview of ALMA. Predictive supervision and LLM-assisted task generation train the contextual flight policy, while only the policy is retained for deployment.}
    \label{fig:overview}
\end{figure*}

\section{Method}
\label{sec:method}

As illustrated in Fig.~\ref{fig:overview}, ALMA addresses the formulation in Sec.~\ref{sec:problem_formulation} through policy representation and training experience, using a contextual flight policy that adapts to the observed system response and predictive supervision that constrains the information retained in its context. Proactive task generation complements this representation learning by using execution feedback to construct additional manipulation tasks.

\subsection{Contextual Flight Policy}
\label{sec:contextual_policy}

ALMA implements the contextual flight policy $\pi_{\theta}(\mathbf{a}_t\mid\mathcal{H}_t)$ through a recurrent architecture operating at control period $\Delta t$, with its observations, action space, and reward specified below.

\paragraph{Architecture}

At each step, the policy receives $\mathbf{o}_t=[\mathbf{r}_t,\mathbf{x}_t,\mathbf{a}_{t-1}]$, comprising the flight reference preview, current measurements, and previous action. A convolutional neural network (CNN) encodes the ordered preview points into the reference feature $\mathbf{z}_t$, while separate multilayer perceptrons (MLPs), termed the state and action encoders, map $\mathbf{x}_t$ to the measurement feature $\mathbf{e}_t$ and encode $\mathbf{a}_{t-1}$, respectively. The measurement and action features are concatenated and passed to a gated recurrent unit (GRU), whose updated hidden state $\mathbf{h}_t$ serves as the policy context and is concatenated with $\mathbf{z}_t$ and $\mathbf{e}_t$ to parameterize the action distribution. Current measurement features provide direct feedback, while the context summarizes recent actions and measured responses, enabling adaptation during deployment through updates from $\mathbf{h}_{t-1}$ to $\mathbf{h}_t$ with fixed network parameters.

\paragraph{Observation}

Let $\mathcal{K}_r$ index the preview points, with $\delta_k$ denoting the look-ahead time of point $k$. At control time $t\Delta t$, the preview $\mathbf{r}_t$ stacks body-frame position errors $\mathbf{e}^{b}_{p,t,k}$ and yaw encodings $(\sin e^{w}_{\psi,t,k},\cos e^{w}_{\psi,t,k})$ between the flight reference at $t\Delta t+\delta_k$ and the latest vehicle pose, with yaw errors computed from world-frame headings. The measurement vector is $\mathbf{x}_t=[\mathbf{v}^{b}_t,\boldsymbol{\omega}^{b}_t,\mathbf{g}^{b}_t,\mathbf{q}_t]$, where $\mathbf{v}^{b}_t$ and $\boldsymbol{\omega}^{b}_t$ are body-frame linear and angular velocities, $\mathbf{g}^{b}_t$ is the gravity vector expressed in the body frame, and $\mathbf{q}_t$ contains arm joint positions, with all observation inputs available at deployment.

\paragraph{Action space}

The action space is $\mathcal{A}=[0,1]^4$, with each component specifying a normalized rotor command mapped to thrust through the actuator model and motor dynamics. Higher-level interfaces such as the collective-thrust and body-rate commands in Aco2~\cite{jin2026autonomous} leave rotor commands to downstream control, and a nominal allocator converts requested thrust and moments using fixed rotor geometry~\cite{Meier_PX4_Autopilot}. For unchanged rotor thrusts, a center-of-mass shift from $\mathbf{c}_0$ to $\mathbf{c}_t$ changes the resultant moment by $\Delta\boldsymbol{\tau}_t=(\mathbf{c}_0-\mathbf{c}_t)\times\mathbf{F}^{\mathrm{rot}}_t$, where $\mathbf{F}^{\mathrm{rot}}_t$ is the resultant rotor force and all vectors are body-frame quantities. Because an allocator defined about $\mathbf{c}_0$ retains the original moment arms while arm inertia and reaction forces and moments additionally change the response, ALMA places individual selection of rotor commands inside the contextual flight policy, allowing $\mathbf{z}_t$, $\mathbf{e}_t$, and $\mathbf{h}_t$ to determine the correction without a separate nominal allocator.

\paragraph{Reward}

The reward $r_t=\Delta t(r_{\mathrm{pose}}-r_{\Delta a}-r_{\mathrm{crash}})$ balances tracking accuracy, control smoothness, and flight safety, with the pose term rewarding position and yaw accuracy and the remaining terms penalizing changes between consecutive actions and crash termination.

\subsection{Predictive Context Learning}
\label{sec:predictive_context}

Motor lag and arm-induced reactions can cause different responses to the same rotor commands despite similar current measurements $\mathbf{x}_t$, making recent measurement--action history relevant to selecting corrections. Control rewards jointly train the representation of this history in $\mathbf{h}_t$ and the actions selected from it, whereas ALMA follows PSD~\cite{venkatraman2017predictive} and RPSP~\cite{hefny2018recurrent} in using subsequent measured responses to give the shared context a direct prediction target.

Following JEPA~\cite{assran2023ijepa}, predictive supervision is defined in a learned representation space, with a predictor estimating future measurement representations from the policy context and recorded actions and a target encoder providing the corresponding targets from subsequent measurements. The state encoder in Sec.~\ref{sec:contextual_policy} is the online measurement encoder shared with the policy, whose features join the action features in the GRU to form $\mathbf{h}_t$. For a prediction horizon of $K$ control steps, the recurrent predictor projects $\mathbf{h}_t$ into $\mathbf{b}_{t,0}$, sequentially processes the recorded rotor commands $\mathbf{a}_t,\ldots,\mathbf{a}_{t+k-1}$, and maps $\mathbf{b}_{t,k}$ to $\widehat{\mathbf{e}}_{t+k}$ through a shared head for $k=1,\ldots,K$, following PBL~\cite{guo2020bootstrap} and the recurrent design of ADM-v2~\cite{lin2026adm}.

The target encoder has the same architecture as the state encoder and maps the recorded $\mathbf{x}_{t+k}$ to $\overline{\mathbf{e}}_{t+k}$. Following SPR~\cite{schwarzer2020data}, we stop gradients through the target branch and update its parameters $\overline{\theta}_s$ by an exponential moving average (EMA) of the parameters of the state encoder $\theta_s$ after each complete policy update, $\overline{\theta}_s\leftarrow\beta\overline{\theta}_s+(1-\beta)\theta_s$, where $0\leq\beta<1$ is the decay. The prediction loss updates the predictor, GRU, and shared state and action encoders, while EMA supplies a gradually updated target representation. For valid same-episode starting indices $\mathcal{I}_k$, the cosine prediction loss is
\begin{equation}
\mathcal{L}_{\mathrm{pred}}
=\frac{1}{K}\sum_{k=1}^{K}\frac{1}{|\mathcal{I}_k|}
\sum_{t\in\mathcal{I}_k}
\left[1-\cos\left(\widehat{\mathbf{e}}_{t+k},\overline{\mathbf{e}}_{t+k}\right)\right].
\label{eq:predictive_loss}
\end{equation}
The prediction target is a learned measurement representation, following JEPA~\cite{assran2023ijepa}, so predictive supervision acts in the feature space that the policy learns for control. We measure prediction error using cosine distance as in SPR~\cite{schwarzer2020data}, which equals one half of the squared Euclidean distance between $\ell_2$-normalized prediction and target vectors. This aligns their directions independently of their overall magnitudes, while averaging over valid samples at each offset and then over the $K$ offsets gives each prediction step equal weight.

Prior analyses of self-predictive learning address optimization dynamics~\cite{tang2023understanding} and the predictive relationship between state and history representations~\cite{ni2024bridging}. To characterize how the prediction objective constrains ALMA's history representation, fix the policy, recurrent initialization, target encoder, trajectory distribution, and offset $k$, and let $H$ be the measurement--action history, with $\mathbf{h}_t=f(H)$, $U=(\mathbf{a}_t,\ldots,\mathbf{a}_{t+k-1})$, and $Y=\overline{\mathbf{e}}_{t+k}/\|\overline{\mathbf{e}}_{t+k}\|_2$ for nonzero targets. Define $\boldsymbol\mu_H=\mathbb{E}[Y\mid H,U]$ and $\boldsymbol\mu_h=\mathbb{E}[Y\mid\mathbf{h}_t,U]$, with unit directions $\mathbf d_H,\mathbf d_h$ chosen arbitrarily for zero means. The minimum expected cosine losses over unrestricted unit-vector predictors are $\ell_j^\star=1-\mathbb{E}\|\boldsymbol\mu_j\|_2$ for $j\in\{H,h\}$, so their difference measures the prediction cost of replacing the full history with the context:
\begin{equation}
\ell_h^\star-\ell_H^\star
=\frac12\mathbb{E}\!\left[
\|\boldsymbol\mu_H\|_2\|\mathbf d_H-\mathbf d_h\|_2^2
\right]\geq0.
\label{eq:context_prediction_gap}
\end{equation}
Indeed, the tower property gives $\mathbb{E}[\mathbf d_h^\top\boldsymbol\mu_H]=\mathbb{E}\|\boldsymbol\mu_h\|_2$, and expanding the squared distance yields the identity. Under the same recorded commands, merging histories that imply different directions of the conditional target means increases the minimum prediction loss. The prediction objective therefore favors a recurrent context that preserves historical distinctions relevant to future system responses.

To relate these distinctions to flight control, consider two trajectories with the same initial vertical velocity, a common fixed near-level attitude, mass $m_{\mathrm{sys}}>0$, motor time constant $\tau_m>0$, and subsequent external-force history, but different initial thrusts. Let $F$ denote the vertical component of rotor thrust, $u$ its steady-state value under a held command, and $\delta$ the difference between trajectories, with $\delta F_0=\delta F(0)$. In a local first-order approximation, $\tau_m\delta\dot F=\delta u-\delta F$ and $m_{\mathrm{sys}}\delta\dot v_z=\delta F$, where $v_z$ is world-frame vertical velocity and $\delta u$ is the difference between constant steady-state thrusts associated with held rotor commands. Solving the thrust equation gives $\delta F(s)=e^{-s/\tau_m}\delta F_0+(1-e^{-s/\tau_m})\delta u$, and integrating the velocity equation over $s\in[0,T]$ for $T>0$ yields
\begin{equation}
m_{\mathrm{sys}}\,\delta v_z(T)
=A_T\,\delta F_0+(T-A_T)\,\delta u,
\label{eq:actuator_response_gap}
\end{equation}
where $A_T=\tau_m(1-e^{-T/\tau_m})$ satisfies $0<A_T<T$. Under a common command, $\delta u=0$, different initial thrusts produce different future velocities, while equal terminal velocities require $\delta u=-A_T\delta F_0/(T-A_T)$ when the corresponding commands are feasible.

When recent measurements and commands distinguish these initial thrusts and the target encoder preserves their velocity effects as different target directions,~\eqref{eq:context_prediction_gap} gives a prediction incentive to retain information needed for the corrections in~\eqref{eq:actuator_response_gap}. Predictive supervision thus trains a control-relevant aspect of the history representation, while the control objective trains how that representation is used to select rotor commands. Deployment uses the contextual flight policy alone.

\subsection{Proactive Task Generation}
\label{sec:task_evolution}

Because predictive supervision learns from the responses present in training trajectories, proactive task generation proposes and evaluates additional arm motions to identify persistent tracking difficulties, using selected motions to broaden experience for response learning and control. Each task $\zeta=(\mathbf r,\mathbf m)$ specifies a flight-reference trajectory and an external arm command sequence, with task generation varying $\mathbf m$ while retaining the tracking objective.

An LLM agent harness~\cite{ning2026code} provides the software layer for this analysis, organizing tool-based access to execution evidence, evidence-linked notes across proposal rounds~\cite{shinn2023reflexion}, and the hand-off from motion proposals to candidate evaluation, so that the LLM can use the records to propose changes in joint coordination and timing. A motion program $g$ specifies participating joints, directions, and the order of movement and holding events, while bounded parameters $\boldsymbol\eta$ specify amplitudes, durations, and starting times. Given a nominal arm reference $\mathbf m_0$, the generator produces $\mathbf m_{g,\boldsymbol\eta}=\mathcal V^m(\mathbf m_0;g,\boldsymbol\eta)$, with a task family defined by fixing $g$ and $\mathbf m_0$ while varying $\boldsymbol\eta$. Parameter search explores that family's instances, while a new program can introduce previously absent joint combinations and event orders. Each instance supplies joint position commands to the arm controller, while ALMA selects rotor commands from the observed response.

Within each family, Sobol sampling proposes bounded parameters whose physically feasible instances are evaluated with the current policy held fixed, initially on a subset of predefined initial conditions and physical parameters. An episode's tracking fraction is the proportion of its duration for which both position and yaw errors remain within their tolerances, with failed episodes assigned zero. Lower mean tracking fractions identify greater tracking difficulty and receive priority, with ties resolved by decreasing position RMSE and then yaw RMSE. Selected candidates serve as parents for local variations and amplitude reductions before evaluation on the full condition set, and retained motions are paired with compatible base-training references and physical configurations. The retained task population $\mathcal P_n$ at stage $n$ is replaced by eligible selected tasks to form $\mathcal P_{n+1}$, or kept unchanged when none are eligible, with evaluation results informing later proposals.

\subsection{Training Procedure}
\label{sec:training_procedure}

Training randomizes vehicle, arm, and payload masses and centers of mass, rotor thrust and moment coefficients, motor response, and servo gains, together with measurement noise, action delays, and external disturbances~\cite{peng2018sim}. An asymmetric actor--critic architecture~\cite{pinto2017asymmetric} uses the contextual flight policy as actor and an independent MLP as critic, with the latter receiving privileged simulator information and a detached $\mathbf{h}_{t-1}$. Policy learning uses Proximal Policy Optimization (PPO)~\cite{schulman2017proximal} to optimize the control return, with prediction errors from recorded action--response sequences providing an additional gradient to the shared recurrent representation through $\mathcal{L}=\mathcal{L}_{\mathrm{PPO}}+\lambda_{\mathrm{pred}}\mathcal{L}_{\mathrm{pred}}$.

Domain randomization varies physical responses, while task generation varies arm-motion experience through episode sampling from $\rho_n=(1-\alpha_n)\rho_{\mathrm{base}}+\alpha_n\rho_{\mathcal P_n}$, where $\rho_{\mathcal P_n}$ samples retained tasks. The generated-task probability $\alpha_n$ starts at $0.05$ when tasks become available, is capped at $0.20$, and is zero for an empty population.

Tracking fractions and crash rates guide the adjustment of $\alpha_n$, which is reduced when performance on base tasks deteriorates, generated tasks are mastered, or crashes persist without improvement. Otherwise, it remains fixed while generated-task performance improves and increases when performance on base tasks is stable but generated-task performance is not improving. The same feedback determines when to request new tasks, allowing exposure to generated tasks to evolve with the policy while continued base-task sampling maintains training on the original motion conditions.

\section{Experiments}
\label{sec:experiments}

The experiments address four questions. \textbf{Q1:} How does ALMA compare with baselines under static and periodic arm commands? \textbf{Q2:} How do context, predictive supervision, and prediction horizon affect tracking? \textbf{Q3:} How do task generation and proposal strategy affect generalization and base-benchmark performance? \textbf{Q4:} Does the policy trained in simulation maintain tracking accuracy in real-world experiments with fixed network parameters?

\subsection{Experimental Setup}
\label{sec:experimental_setup}

\paragraph{Training and evaluation}

All learned controllers are trained in NVIDIA Isaac Sim~\cite{NVIDIA_Isaac_Sim}. ALMA operates at $100$~Hz, with $K=5$ prediction steps ($10$--$50$~ms), $\lambda_{\mathrm{pred}}=0.1$, and target-encoder EMA decay $\beta=0.85$ after each complete PPO update, while motor time constants are randomized over $30$--$60$~ms. Policy training takes approximately one hour on a single NVIDIA RTX~4090 GPU, with GPT-5.6 Sol task proposals costing approximately \$15 per run. MuJoCo~\cite{todorov2012mujoco} provides the simulated plant in hardware-in-the-loop (HIL) evaluations, with a physical flight controller executing control at the same frequency, while real-world experiments evaluate the same policy trained in simulation on the aerial manipulator with fixed network parameters.

\paragraph{Experimental platform}

The quadrotor has a diagonal wheelbase of $500$~mm, a total takeoff mass of $2.39$~kg, and a maximum total thrust of $98.68$~N, while its four-degree-of-freedom (4-DoF) arm and two-finger parallel gripper together weigh $0.516$~kg and provide a maximum reach of approximately $0.62$~m. The propulsion system uses T-MOTOR F100 KV1350 motors, HQProp HQ-8045-3 propellers, and a $6$S $4600$~mAh battery, with vehicle pose supplied by external motion capture.

\paragraph{Benchmark and evaluation protocol}

\begin{figure*}[htbp]
    \centering
    \includegraphics[width=1.0\linewidth]{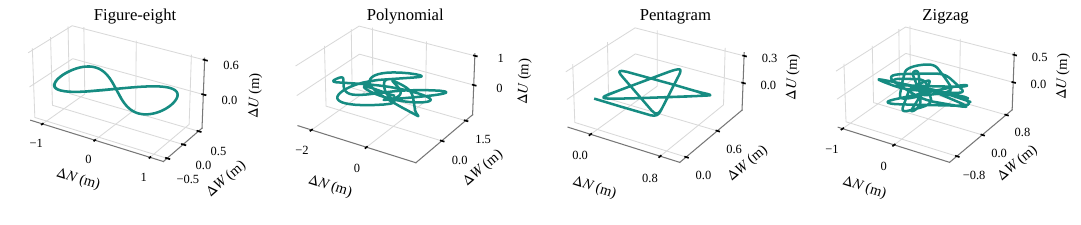}
    \caption{Representative flight-reference trajectories in the main benchmark. From left to right: Figure-eight, Polynomial, Pentagram, and Zigzag. The benchmark also includes Hover (not shown) and evaluates the Figure-eight and Pentagram trajectories at multiple speeds.}
    \label{fig:benchmark_trajectories}
\end{figure*}

The main benchmark combines Hover with the Figure-eight, Polynomial, Pentagram, and Zigzag references adapted from SimpleFlight~\cite{chen2025simpleflight} (Fig.~\ref{fig:benchmark_trajectories}). Controllers track identical flight references under identical external arm commands that hold selected configurations in static tests and repeat prescribed movements in periodic tests, while each baseline retains its native observation and control interface. Position RMSE and yaw RMSE are computed over successful trials using the three-dimensional Euclidean position error and wrapped yaw error, respectively, while success rate (SR) is the fraction of successful trials among all attempts. Learned controllers use three training seeds, with aggregation specified in each table note.

\subsection{Flight Tracking}
\label{sec:tracking_results}

For \textbf{Q1}, we compare ALMA with five baselines: \textbf{PID}~\cite{Meier_PX4_Autopilot} uses the PX4 cascaded controller with fixed gains and nominal allocation, while \textbf{PID--INDI}~\cite{smeur2016adaptive} retains the PID outer loops and uses an INDI inner loop with angular velocity, angular acceleration, and rotor-speed feedback. \textbf{PID}+$\mathcal{L}_1$~\cite{wu20251} adds filtered translational and rotational corrections, while our \textbf{NMPC} implementation incorporates manipulator dynamics into the finite-horizon prediction model~\cite{sun2022comparative}. The adapted controller of \textbf{Zhang et al.}~\cite{zhang2025extreme} infers latent dynamics from recent measurements and actions and additionally receives arm observations.

\begin{table*}[htbp]
    \centering
    \caption{Overall Flight Tracking Performance}
    \label{tab:main_results}

    \small
    \setlength{\tabcolsep}{2.5pt}
    \renewcommand{\arraystretch}{1.16}

    \begin{tabular*}{1.0\textwidth}{
        @{\extracolsep{\fill}}
        l
        ccc
        ccc
        @{}
    }
        \toprule
        &
        \multicolumn{3}{c}{Static Arm Commands}
        &
        \multicolumn{3}{c}{Periodic Arm Commands}
        \\
        \cmidrule(lr){2-4}
        \cmidrule(lr){5-7}

        Method
        & \shortstack{Position RMSE\\(cm) $\downarrow$}
        & \shortstack{Yaw RMSE\\($^\circ$) $\downarrow$}
        & \shortstack{SR\\(\%) $\uparrow$}
        & \shortstack{Position RMSE\\(cm) $\downarrow$}
        & \shortstack{Yaw RMSE\\($^\circ$) $\downarrow$}
        & \shortstack{SR\\(\%) $\uparrow$}
        \\
        \midrule

        PID~\cite{Meier_PX4_Autopilot}
        & $4.514 \pm 0.180$
        & $0.299 \pm 0.178$
        & $100$
        & $42.479 \pm 15.746$
        & $0.577 \pm 0.178$
        & $100$
        \\

        PID--INDI~\cite{smeur2016adaptive}
        & $4.230 \pm 0.036$
        & $\mathbf{0.031 \pm 0.006}$
        & $100$
        & $6.770 \pm 2.041$
        & $\mathbf{0.037 \pm 0.016}$
        & $100$
        \\

        PID+$\mathcal{L}_1$~\cite{wu20251}
        & $3.893 \pm 0.162$
        & $0.263 \pm 0.179$
        & $100$
        & $24.654 \pm 6.896$
        & $0.426 \pm 0.166$
        & $100$
        \\

        NMPC~\cite{sun2022comparative}
        & $\underline{3.082 \pm 0.618}$
        & $\underline{0.180 \pm 0.078}$
        & $100$
        & $\underline{3.315 \pm 0.089}$
        & $\underline{0.176 \pm 0.096}$
        & $100$
        \\

        \midrule

        Zhang et al.~\cite{zhang2025extreme}
        & $6.434 \pm 2.473$
        & $1.279 \pm 1.279$
        & $83.33$
        & $10.794 \pm 2.689$
        & $2.459 \pm 1.221$
        & $100$
        \\

        \textbf{ALMA (Ours)}
        & $\mathbf{2.326 \pm 0.332}$
        & $0.439 \pm 0.132$
        & $100$
        & $\mathbf{2.497 \pm 0.209}$
        & $0.499 \pm 0.089$
        & $100$
        \\

        \bottomrule
    \end{tabular*}

    \vspace{2pt}
    \begin{minipage}{\textwidth}
        \footnotesize
        \raggedright
        \textit{Note:}
        The upper and lower blocks contain conventional and learning-based controllers, respectively. Flight-reference results are weighted equally within each arm command condition, with results for learning-based controllers first averaged across three training seeds. Mean $\pm$ sample standard deviation is then reported across arm command conditions, while SR includes all attempted trials. The lowest and second-lowest mean RMSE values are shown in bold and underlined, respectively.
    \end{minipage}
\end{table*}

Table~\ref{tab:main_results} shows that ALMA achieves the lowest mean position RMSE under both static and periodic arm commands, reducing error relative to NMPC, the strongest position-tracking baseline, by $24.5\%$ and $24.7\%$, respectively, while PID--INDI achieves the lowest yaw RMSE in both settings. Relative to Zhang et al.~\cite{zhang2025extreme}, ALMA reduces position and yaw RMSE under periodic arm commands by $76.9\%$ and $79.7\%$, respectively, with an SR of $100\%$ in both settings compared with that baseline's $83.33\%$ under static commands.

ALMA maintains its position-tracking advantage under periodic arm commands, with position RMSE increasing by $7.4\%$ from the static setting, close to NMPC's $7.6\%$. Static configurations change the mass distribution and inertia, whereas periodic motion additionally produces time-varying reaction forces and moments, making the recent motion history relevant to the required correction. Within ALMA, the contextual flight policy combines current measurements with recent actions and measured responses to select rotor commands, while the following ablation in Sec.~\ref{sec:context_results} examines how predictive supervision contributes to learning this history representation.

\subsection{Predictive Context Learning}
\label{sec:context_results}

\begin{table}[t]
    \centering
    \caption{Ablation Study of Predictive Context Learning}
    \label{tab:context_ablation}
    \small
    \setlength{\tabcolsep}{5pt}
    \renewcommand{\arraystretch}{1.13}

    \begin{tabular*}{\columnwidth}{
        @{\extracolsep{\fill}}
        lcc
        @{}
    }
        \toprule
        Variant
        & \shortstack{Position RMSE\\(cm) $\downarrow$}
        & \shortstack{Yaw RMSE\\($^\circ$) $\downarrow$}
        \\
        \midrule

        \textsc{Feedforward}
        & $3.982 \pm 0.196$
        & $0.798 \pm 0.036$
        \\

        \textsc{GRU Only}
        & $4.020 \pm 0.153$
        & $0.647 \pm 0.092$
        \\

        \textsc{Pred-1}
        & $3.989 \pm 0.185$
        & $\mathbf{0.474 \pm 0.033}$
        \\

        \textsc{Pred-3}
        & $3.872 \pm 0.319$
        & $0.538 \pm 0.101$
        \\

        \textbf{\textsc{Pred-5} (Ours)}
        & $\mathbf{3.576 \pm 0.167}$
        & $\underline{0.485 \pm 0.045}$
        \\

        \textsc{Pred-10}
        & $\underline{3.824 \pm 0.157}$
        & $0.545 \pm 0.028$
        \\

        \bottomrule
    \end{tabular*}

    \par\vspace{0.8mm}
    \begin{minipage}{\columnwidth}
        \footnotesize
        \raggedright
        \textit{Note:}
            Results are averaged equally across evaluation conditions within each seed and reported as mean $\pm$ sample standard deviation across training seeds. Best and second best results are shown in bold and underlined, respectively. Task generation is disabled for all variants.
    \end{minipage}
\end{table}

For \textbf{Q2}, the benchmark adds payload changes and transient and sustained external forces that alter the response even at similar arm configurations. \textsc{Feedforward} replaces the GRU with a parameter-matched MLP, while \textsc{GRU Only} uses recurrence without predictive supervision. With task generation disabled throughout this ablation, \textsc{Pred-1}, \textsc{Pred-3}, \textsc{Pred-5}, and \textsc{Pred-10} add prediction targets at steps $1,\ldots,K$ for their respective horizons.

Table~\ref{tab:context_ablation} shows lower yaw RMSE for \textsc{GRU Only} than \textsc{Feedforward}, with similar position means, while \textsc{Pred-5} further reduces position and yaw RMSE by $11.0\%$ and $25.0\%$ relative to \textsc{GRU Only}. Since the recurrent actor architecture is unchanged, this comparison isolates the effect of training its context with future-response prediction, connecting the representation constraint in~\eqref{eq:context_prediction_gap} to improved tracking performance.

\textsc{Pred-5} has the lowest mean position RMSE, with its five-step targets covering more of the transient described by a first-order motor model, in which the completed fraction of a step response at time $T$ is $1-\exp(-T/\tau_m)$. For $\tau_m=30$--$60$~ms, this fraction is $15$--$28\%$ at $10$~ms and $57$--$81\%$ at $50$~ms, providing a possible mechanism for improved position tracking over the five-step horizon. Although \textsc{Pred-1} has the lowest mean yaw RMSE, $0.011^\circ$ lower than that of \textsc{Pred-5}, we choose $K=5$ for its position-tracking performance.

\textsc{Pred-10} has higher mean position and yaw RMSE than \textsc{Pred-5}, while equal weighting in~\eqref{eq:predictive_loss} halves the combined weight of the first five targets when $K$ doubles. Because more distant targets can also depend more strongly on future arm commands absent from the predictor input, the resulting shift in supervision may reduce the emphasis on immediate, predictable responses.

\subsection{Proactive Task Generation}
\label{sec:task_generation_results}

For \textbf{Q3}, we use Hover to evaluate generalization across different arm motions while holding the flight reference fixed, allowing the comparison to focus on tracking under arm-induced responses. The evaluation covers out-of-distribution (OOD) arm motions with joint velocities and accelerations beyond the training ranges, reserved family variants with unseen parameter settings, and the base benchmark. \textsc{ALMA-Base} retains the contextual policy and predictive supervision but trains only on $\rho_{\mathrm{base}}$, while \textsc{Random Evolution} replaces LLM proposals with random structural variations under the same search and selection procedure.

\begin{table*}[htbp]
    \centering
    \caption{Ablation of Proactive Task Generation}
    \label{tab:task_generation}

    \small
    \setlength{\tabcolsep}{3pt}
    \renewcommand{\arraystretch}{1.16}

    \begin{tabular*}{\textwidth}{
        @{\extracolsep{\fill}}
        l
        *{6}{c}
        @{}
    }
        \toprule
        &
        \multicolumn{2}{c}{OOD Arm Motions}
        &
        \multicolumn{2}{c}{Reserved Family Variants}
        &
        \multicolumn{2}{c}{Base Benchmark}
        \\
        \cmidrule(lr){2-3}
        \cmidrule(lr){4-5}
        \cmidrule(lr){6-7}

        Method
        & \shortstack{Position RMSE\\(cm) $\downarrow$}
        & \shortstack{Yaw RMSE\\($^\circ$) $\downarrow$}
        & \shortstack{Position RMSE\\(cm) $\downarrow$}
        & \shortstack{Yaw RMSE\\($^\circ$) $\downarrow$}
        & \shortstack{Position RMSE\\(cm) $\downarrow$}
        & \shortstack{Yaw RMSE\\($^\circ$) $\downarrow$}
        \\
        \midrule

        \textsc{ALMA-Base}
        & $2.937 \pm 0.313$
        & $\underline{1.048 \pm 0.291}$
        & $5.839 \pm 3.949$
        & $6.322 \pm 3.523$
        & $\underline{2.326 \pm 0.303}$
        & $0.641 \pm 0.066$
        \\

        \textsc{Random Evolution}
        & $\underline{2.807 \pm 0.425}$
        & $1.211 \pm 0.439$
        & $\underline{5.190 \pm 2.486}$
        & $\mathbf{5.543 \pm 2.587}$
        & $2.354 \pm 0.205$
        & $\underline{0.586 \pm 0.048}$
        \\

        \textbf{ALMA (Ours)}
        & $\mathbf{2.490 \pm 0.343}$
        & $\mathbf{0.935 \pm 0.246}$
        & $\mathbf{3.803 \pm 1.439}$
        & $\underline{5.726 \pm 2.692}$
        & $\mathbf{2.148 \pm 0.332}$
        & $\mathbf{0.477 \pm 0.049}$
        \\

        \bottomrule
    \end{tabular*}

    \par\vspace{0.8mm}
    \begin{minipage}{\textwidth}
        \footnotesize
        \raggedright
        \textit{Note:}
            \textsc{ALMA-Base} trains only on $\rho_{\mathrm{base}}$, while \textsc{Random Evolution} replaces ALMA's LLM proposals with random structural variations while retaining the same evolution and selection procedure. Within each training seed, evaluation conditions are weighted equally within each benchmark. Entries report the mean $\pm$ sample standard deviation across training seeds. Bold and underlined values indicate the lowest and second-lowest mean RMSE in each column, respectively.
    \end{minipage}
\end{table*}

Table~\ref{tab:task_generation} reveals different patterns of generalization across the evaluated motion conditions. Relative to \textsc{ALMA-Base}, \textsc{Random Evolution} improves four of the six mean RMSE metrics, but its $12.3\%$ reduction in yaw RMSE on reserved family variants is accompanied by a $15.6\%$ increase in OOD yaw error. In contrast, ALMA improves all six metrics, achieving a further $22.8\%$ reduction in OOD yaw RMSE relative to \textsc{Random Evolution}. This difference highlights the role of motion-structure proposals in generalization: since both methods use the same candidate search and selection procedure, their different generalization outcomes suggest that the proposed motion structures influence transfer, as parameter search can refine a given structure but cannot introduce new joint coordination patterns or event sequences.

\subsection{Real-World Experiments}
\label{sec:real_world}

For \textbf{Q4}, the policy trained in simulation is deployed on the onboard flight controller with fixed network parameters and compared with PID in real-world experiments. Each control step uses one forward pass of the contextual flight policy, with inference taking $1.715\pm0.030$~ms, approximately $17\%$ of the $10$~ms control period.

\begin{figure*}[htbp]
    \centering
    \includegraphics[width=1.0\linewidth]{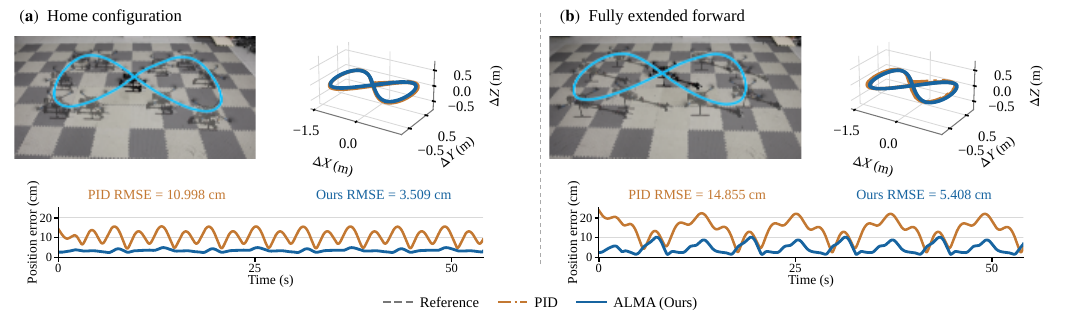}
    \caption{Zero-shot trajectory tracking in real-world experiments with the arm in (a) the home configuration and (b) the fully extended forward configuration.}
    \label{fig:real-world}
\end{figure*}

\begin{figure}[htbp]
    \centering
    \includegraphics[width=0.5\linewidth]{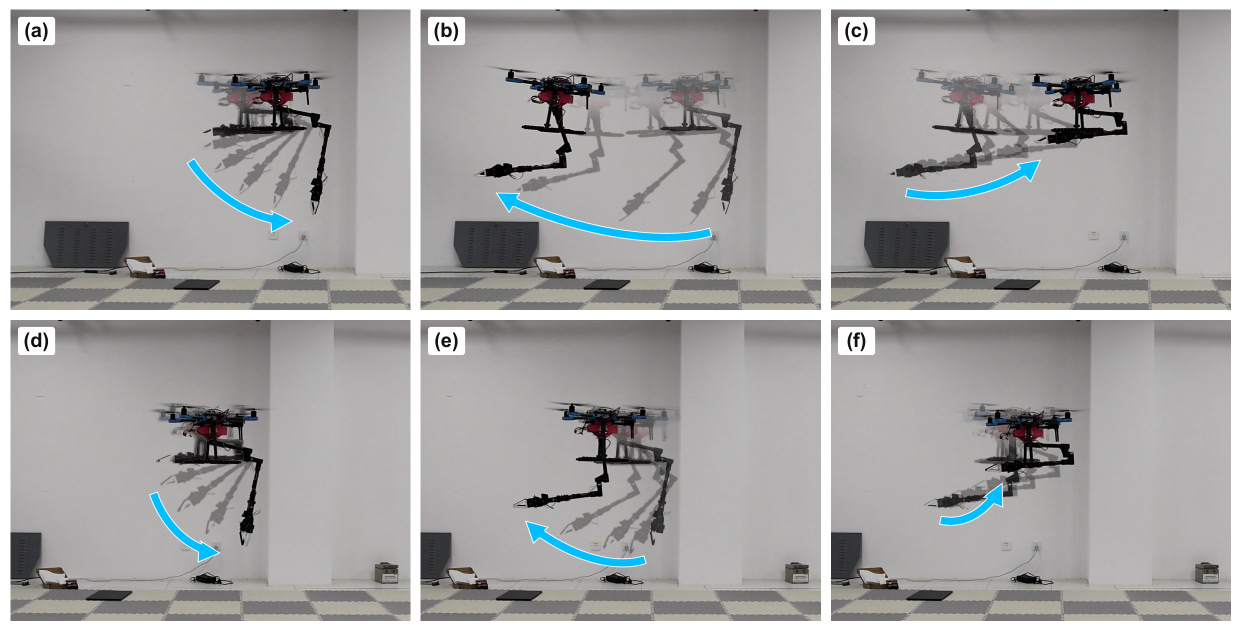}
    \caption{Real-world hover stabilization under periodic arm commands. Panels (a)--(c) show PID and panels (d)--(f) show ALMA at corresponding stages of the same arm motion. Position RMSE is $3.17$~cm for ALMA and $15.55$~cm for PID.}
    \label{fig:hover}
\end{figure}

The same policy has lower position RMSE on the Figure-eight reference and a reference tracing the letters ICRA in the home and fully extended forward configurations (Fig.~\ref{fig:real-world}). As extension changes the center of mass and inertia, joint-position feedback indicates the configuration while recent measurements and rotor commands describe the response under that configuration, allowing these experiments to evaluate their combined use by one contextual flight policy.

During hover under periodic arm commands with an $8$~s arm-motion period, ALMA reduces position RMSE from PID's $15.55$~cm to $3.17$~cm, a $79.6\%$ reduction (Fig.~\ref{fig:hover}). Together, the trajectory-tracking and hover experiments demonstrate zero-shot transfer of a single contextual flight policy across fixed arm configurations and continuous arm motion.

\section{Conclusions}
\label{sec:conclusions}

This paper presented \textbf{ALMA} for adaptive flight control under external arm commands, combining action-conditioned predictive supervision of the policy context with LLM-assisted task generation guided by execution feedback. The analysis relates historical response information to actuator-dependent corrections, while simulation results show the lowest mean position RMSE among evaluated controllers, with up to a \textbf{24.7\%} reduction relative to NMPC. Ablations support the tracking benefit of predictive supervision and the OOD generalization benefit of task generation, while real-world experiments demonstrate zero-shot deployment with a \textbf{79.6\%} reduction in hover position RMSE relative to PID during periodic arm motion. Future work will investigate improved yaw tracking and contact-aware control with force feedback and contact constraints.

\bibliographystyle{unsrtnat}
\bibliography{references}  %%% Uncomment this line and comment out the ``thebibliography'' section below to use the external .bib file (using bibtex) .




% 	\bibitem{keshet2016prediction}
% 	Keshet, Renato, Alina Maor, and George Kour.
% 	\newblock Prediction-Based, Prioritized Market-Share Insight Extraction.
% 	\newblock In {\em Advanced Data Mining and Applications (ADMA), 2016 12th International 
%                       Conference of}, pages 81--94,2016.

% \end{thebibliography}

\end{document}